# Exploring compact reinforcement-learning representations with linear regression

**Thomas J. Walsh**[†]   **István Szita**[‡]   **Carlos Diuk**[†]   **Michael L. Littman**[†]

[†]Dept. of Computer Science
Rutgers University
Piscataway, NJ 08854

[‡]Dept. of Computing Science
University of Alberta
Edmonton, AB Canada T6G 2E8

## Abstract

This paper presents a new algorithm for online linear regression whose efficiency guarantees satisfy the requirements of the KWIK (Knows What It Knows) framework. The algorithm improves on the complexity bounds of the current state-of-the-art procedure in this setting. We explore several applications of this algorithm for learning compact reinforcement-learning representations. We show that KWIK linear regression can be used to learn the reward function of a factored MDP and the probabilities of action outcomes in Stochastic STRIPS and Object Oriented MDPs, none of which have been proven to be efficiently learnable in the RL setting before. We also combine KWIK linear regression with other KWIK learners to learn larger portions of these models, including experiments on learning factored MDP transition and reward functions together.

## 1 Introduction

Linear regression has, for decades, been a powerful tool in the kits of machine-learning researchers. While the field of Reinforcement Learning (RL) [16] has certainly made use of linear regression in approximating value functions [3], using online regression to learn parameters of a model has been limited to environments with linear dynamics (e.g. [7]), and has often been unable to make guarantees about the behavior of the resulting learning agent without strict assumptions. One of the great hindrances in applying linear regression to learn models in any RL environment was that the computational and sample efficiency guarantees of online regression learners (such as those that rely on distributional assumptions or do not maintain explicit confidences) did not port to the reinforcement-learning setting, which is not i.i.d. and where realizing what portion of the model needs to be explored is crucial to optimizing reward.

Recently, the introduction of the KWIK (Knows What It Knows) framework [11] has provided a characterization of sufficient conditions for a model-learning algorithm to induce sample-efficient behavior in a reinforcement-learning agent. One of the first algorithms developed for this framework was a KWIK linear regression algorithm [15], which was used to learn the transition function of an MDP with linear dynamics. In this paper, we present an algorithm that improves on both the sample and computational bounds of this previous algorithm and apply it to a stable of learning problems for reinforcement-learning agents that employ "compact" representations. Specifically, we use KWIK linear regression (KWIK-LR) to learn the reward function in a factored MDP and the transition probabilities in domains encoded using Stochastic STRIPS [12] or Object Oriented MDPs (OOMDP) [5]. We note that learning these parameters is not typically associated with linear regression—this paper shows that KWIK-LR can be used to help learn models beyond its standard usage in learning linear dynamics. Because of the KWIK guarantees, agents using this algorithm in these settings are guaranteed to make at most a polynomial (in the parameters of the learned model) number of sub-optimal steps with high probability. We present algorithms and theoretical arguments to this effect, including a general reinforcement-learning algorithm for agents that need to learn the probabilities of action outcomes when these effects may be ambiguous in sample data. Experimental evidence is also presented for benchmark problems in the application areas mentioned above.

Our major contributions are an improved and simplified algorithm for KWIK linear regression, and the sample-efficient algorithms that use KWIK-LR to efficiently learn portions of compact reinforcement-learning representations, none of which have previ-ously been shown to be efficiently learnable. We also discuss extensions, including combining KWIK-LR with other KWIK algorithms to learn more parameters of these compact representations.

## 2 KWIK Linear Regression

KWIK [11] (Knows What It Knows) is a framework for studying supervised learning algorithms and was designed to unify the analysis of model-based reinforcement-learning algorithms. Formally, a KWIK learner operates over an input space $X$ and an output space $Y$. At every timestep $t$, an input $x_t \in X$ is chosen and presented to the learner. If the learner can make an accurate prediction on this input, it can predict $\widehat{y}_t$, otherwise it must admit it does not know by returning $\perp$ ("I don't know"), allowing it to see the true $y_t$ or a noisy version $z_t$. An algorithm is said to be KWIK if and only if, with high $(1-\delta)$ probability, $\|\widehat{y}_t - y_t\| < \epsilon$ and the number of $\perp$s returned over the agent's lifetime is bounded by a polynomial function over the size of the input problem. It has been shown [10] that in the model-based reinforcement-learning setting, if the underlying model learner is KWIK, then it is possible to build an RL agent around it by driving exploration of the $\perp$ areas using an R-max [4] style manipulation of the value function. Such an agent will, with high probability, take no more than a polynomial (in the parameters of the model being learned) number of suboptimal actions. In this paper, which deals with compact representations where the parameter sizes are far smaller than the enumerated state space, we will call agents that satisfy these conditions, but may require super-polynomial (in the size of the compact parameters) computation for planning, *PAC-compact-MDP*.

One of the first uses of the KWIK framework was in the analysis of an online linear regression algorithm used to learn linear transitions in continuous state MDPs [15]. This algorithm uses the least squares estimate of the weight vector for inputs where the output is known with high certainty. Certainty is measured by two terms representing (1) the number and proximity of previous samples to the current point and (2) the appropriateness of the previous samples for making a least squares estimate. When certainty is low for either measure, the algorithm reports $\perp$. In this work, we present a KWIK-LR algorithm that is more sample and computationally efficient than this previous work. First though, we define some notation.

Let $X := \{\vec{x} \in \mathbb{R}^n \mid \|\vec{x}\| \leq 1\}$, and let $f : X \to \mathbb{R}$ be a linear function with slope $\theta^* \in \mathbb{R}^n$, $\|\theta^*\| \leq M$, i.e. $f(\vec{x}) := \vec{x}^T \theta^*$. Fix a timestep $t$. For each $i \in \{1, \ldots, t\}$, denote the stored samples by $\vec{x}_i$, their (unknown) expected values by $y_i := \vec{x}_i^T \theta^*$, and their observed values by $z_i := \vec{x}_i^T \theta^* + \eta_i$, where the noise $\eta_i$ is assumed to form a martingale, i.e., $E(\eta_i | \eta_1, \ldots, \eta_{i-1}) = 0$, and bounded: $|\eta_i| \leq S$. Define the matrix $D_t := [\vec{x}_1, \vec{x}_2, \ldots, \vec{x}_t]^T \in \mathbb{R}^{t \times n}$ and vectors $\vec{y}_t := [y_1; \ldots; y_t] \in \mathbb{R}^t$ and $\vec{z}_t := [z_1; \ldots; z_t] \in \mathbb{R}^t$, and let $I$ be an $n \times n$ identity matrix.

### 2.1 A New KWIK Linear Regression Algorithm

Suppose that a new query $\vec{x}$ arrives. If we were able to solve the linear regression problem $D_t \theta = \vec{z}_t$, then we could predict $\widehat{y} = \vec{x}^T \theta$, where $\theta$ is the least-squares solution to the system. However, solving this system directly is problematic because: (1) if $D_t$ is rank-deficient the least-squares solution may not be unique and (2) even if we have a solution, we have no information on its confidence.

We can avoid the first problem by regularization, i.e. by augmenting the system with $I\theta = \vec{v}$, where $\vec{v}$ is some arbitrary vector. Regularization certainly distorts the solution, but this gives us a measure of confidence: if the distortion is large, the predictor should have low confidence and output $\perp$. On the other hand, if the distortion is low, it has two important consequences. First, the choice of $\vec{v}$ has little effect, and second, the fluctuations caused by using $\vec{z}_t$ instead of $\vec{y}_t$ are also minor.

Let $A_t := [I; D_t^T]^T$. The solution of the system $A_t \theta = [(\theta^*)^T; \vec{y}_t^T]^T$ is unique, and equal to $\theta^*$. However, the right-hand side of this system includes the unknown $\theta^*$, so we use the approximate system $A_t \theta = [\vec{0}^T; \vec{z}_t^T]^T$, which has a solution $\widehat{\theta} = (A_t^T A_t)^{-1} A_t^T [\vec{0}^T; \vec{z}_t^T]^T$. Define $Q_t := (A_t^T A_t)^{-1}$. The prediction error for $\vec{x}$ is

$$\widehat{y} - y = \vec{x}^T(\widehat{\theta} - \theta^*) \quad (1)$$
$$= \vec{x}^T Q_t A_t^T \left( \begin{bmatrix} \vec{0} \\ \vec{z}_t \end{bmatrix} - \begin{bmatrix} \theta^* \\ \vec{y}_t \end{bmatrix} \right) = \vec{x}^T Q_t A_t^T \left( \begin{bmatrix} \vec{0} \\ \vec{\eta}_t \end{bmatrix} - \begin{bmatrix} \theta^* \\ \vec{0} \end{bmatrix} \right).$$

If $\|Q_t \vec{x}\|$ is small, then both terms of (1) will be small, although for different reasons. If $\|Q_t \vec{x}\|$ is larger than a suitable threshold $\alpha_0$, our algorithm will output $\perp$, otherwise it outputs $\widehat{y}$, which will be an accurate prediction with high probability.

Algorithm 1 describes our method for KWIK-learning a linear model. Notice it avoids the problem of storing $A_t$ and $\vec{z}_t$, which grow without bound as $t \to \infty$. The quantities $Q_t = (A_t^T A_t)^{-1}$ and $\vec{w}_t = A_t^T [\vec{0}^T, \vec{z}_t^T]^T$ are sufficient for calculating the predictions, and can be updated incrementally (see Algorithm 1). The algorithm is KWIK, as shown next.

**Theorem 2.1** *Let $\delta > 0$ and $\epsilon > 0$. If Algorithm 1 is executed with $\alpha_0 := \min\left\{\frac{c \cdot \epsilon^2}{\log \frac{n}{\delta \epsilon}}, \frac{\epsilon}{2M}\right\}$ with* a suitable constant $c$, then the number of $\perp$s will be $O\left(\frac{n}{\epsilon^2}\max\left\{\frac{1}{\epsilon^2}\log^2\frac{n}{\delta\epsilon}, M^2\right\}\right)$, and with probability at least $1 - \delta$, for each sample $\vec{x}_t$ for which a prediction $\widehat{y}_t$ is made, $|\widehat{y}_t - f(\vec{x}_t)| \leq \epsilon$ holds.[1]

**Algorithm 1** Learning of linear model
  **input:** $\alpha_0$
  **initialize:** $t := 0$, $m := 0$, $Q := I$, $\vec{w} := \vec{0}$
  **repeat**
    observe $\vec{x}_t$
    **if** $\|Q\vec{x}_t\| < \alpha_0$ **then**
      predict $\widehat{y}_t = \vec{x}_t^T Q \vec{w}$ //known state
    **else**
      predict $\widehat{y}_t = \perp$ //unknown state
    observe $z_t$
    $Q := Q - \frac{(Q\vec{x}_t)(Q\vec{x}_t)^T}{1+\vec{x}_t^T Q \vec{x}_t}$, $\vec{w} := \vec{w} + \vec{x}_t z_t$
    $t := t + 1$
  **until** there are more samples

This result is a $\Theta(n^2/\log^2 n)$ improvement over the sample complexity of Strehl & Littman's KWIK online linear regression, and requires $\Theta(n^2)$ operations per timestep, in contrast to their $O(tn^2)$ complexity.

### 2.2 Proof sketch of Theorem 2.1

The second term of the prediction error (1) is $\vec{x}^T Q \theta^*$, which can be bounded by $|\vec{x}^T Q \theta^*| \leq \|Q\vec{x}\| \|\theta^*\| \leq \alpha_0 M$. This will be no more than $\epsilon/2$ if $\alpha_0 \leq \epsilon/(2M)$.

We now consider the first term of (1). Fix a constant $\alpha \in \mathbb{R}$, and let $m$ be the number of timesteps when $\|Q_t \vec{x}_t\| > \alpha$. We show that $m < 2n/\alpha^2$. We proceed by showing that the traces of the $Q_t$ matrices are positive, monotonically decreasing, and decrease considerably when $\|Q_t \vec{x}_t\|$ is large. $Q_1 = I$, so $\operatorname{trace}(Q_1) = n$. For $t \geq 1$, $Q_{t+1} - Q_t = -\frac{(Q_t \vec{x}_t)(Q_t \vec{x}_t)^T}{1+\vec{x}_t^T Q_t \vec{x}_t}$. To lowerbound the update, note that $A_t^T A_t = (I + D_t^T D_t) \geq I$, so $Q_t = (A_t^T A_t)^{-1} \leq I$. Therefore, $1 + \vec{x}_t^T Q_t \vec{x}_t \leq 1 + \vec{x}_t^T \vec{x}_t \leq 2$.

$$\operatorname{trace}(Q_{t+1}) - \operatorname{trace}(Q_t) = -\frac{\operatorname{trace}\left((Q_t\vec{x}_t)(Q_t\vec{x}_t)^T\right)}{1+\vec{x}_t^T Q_t \vec{x}_t}$$
$$\leq -\frac{1}{2}\operatorname{trace}\left((Q_t\vec{x}_t)(Q_t\vec{x}_t)^T\right) = -\frac{1}{2}\|Q_t\vec{x}_t\|^2, \quad (2)$$

When $\|Q_t \vec{x}_t\| > \alpha$, this expression is at most $-\frac{\alpha^2}{2}$ and at most 0 otherwise. Applying (2) iteratively, we get that $\operatorname{trace}(Q_{t+1}) \leq \operatorname{trace}(Q_1) - m\frac{\alpha^2}{2} = n - m\frac{\alpha^2}{2}$. On the other hand, $Q_{t+1}$ is positive definite, which implies that $\operatorname{trace}(Q_{t+1})$ is positive. So, $m < 2n/\alpha^2$.

Let $\{\alpha_k\}$ ($k = 0, 1, \ldots$) be a monotonically decreasing, 0-limit sequence of constants to be defined later, and let $m_k$ be the number of samples where $\|Q_t \vec{x}_t\| > \alpha_k$ is true.

---
[1] The theorem also holds if the noise is Gaussian. There will be a difference only in the constant terms.

Fix $t$ and drop the subscripts of $\vec{x}_t$, $A_t$ and $Q_t$. Suppose a new sample $\vec{x}$ arrives and the algorithm decides that it is "known", i.e., $\|Q\vec{x}\| \leq \alpha_0$. The first term of the prediction error (1) is a weighted sum of the noise, $\sum_{i=1}^t (\vec{x}^T Q \vec{x}_i)\eta_i$. If $\|Q\vec{x}\| = 0$ then the prediction error is 0. Otherwise let $k$ be the index for which $\alpha_{k+1} < \|Q\vec{x}\| \leq \alpha_k$. The number of $\vec{x}$ inputs falling in the range $(\alpha_{k+1}, \alpha_k]$ is at most $m_{k+1}$. We can bound the squared sum of the weights of the noise terms:

$$\sum_{i=1}^t (\vec{x}^T Q \vec{x}_i)^2 = \sum_{i=n+1}^{n+t} \left([\vec{x}^T Q A^T]_i\right)^2 < (AQ\vec{x})^T (AQ\vec{x})$$
$$= \vec{x}^T Q(A^T A Q)\vec{x} = \vec{x}^T Q \vec{x} \leq \|Q\vec{x}\| \|\vec{x}\| \leq \alpha_k.$$

Let $\delta_1 > 0$ be a constant to be determined later. We can apply Azuma's inequality to the weighted sum of random variables $\eta_i$, with weights $\vec{x}^T Q \vec{x}_i$, which gives the result that the probability that the first term of the error is larger than $\epsilon/2$ is at most $2\exp\left(-\frac{\epsilon^2}{16S^2 \alpha_k}\right)$. This probability will be no more than $\delta_1/2^k$ if $\alpha_k \leq \frac{\epsilon^2}{16S^2 \log(2^{k+1}/\delta_1)}$. Putting together the two terms, if $\alpha_k := \min\left\{\frac{\epsilon^2}{16S^2 \log(2^{k+1}/\delta_1)}, \frac{\epsilon}{2M}\right\}$, then $|\widehat{y} - y| < \epsilon$ with probability at least $1 - \delta_1/2^k$. With these settings, the number of $\perp$s is at most $m_0 \leq \frac{2n}{\alpha_0^2} = \Theta\left(\max\left\{\frac{n\log^2(1/\delta_1)}{\epsilon^4}, \frac{nM^2}{\epsilon^2}\right\}\right)$. The total probability of an error is at most $\sum_{k=0}^\infty m_k \frac{\delta_1}{2^k} \leq \sum_{k=0}^\infty \frac{2n}{\alpha_k^2} \frac{\delta_1}{2^k} = \Theta\left(\max\left\{\frac{n\delta_1 \log^2(1/\delta_1)}{\epsilon^4}, \frac{n\delta_1 M^2}{\epsilon^2}\right\}\right)$. Let us set $\delta_1$ so that the above probability is less than $\delta$. For this, the following assignment is sufficient: $\delta_1 := O\left(\min\left\{\frac{\epsilon^4 \delta^2}{n}, \frac{\epsilon^2 \delta}{nM^2}\right\}\right)$.

## 3 Application 1: Learning Rewards in a Factored-State MDP

A Markov decision process (MDP) [16] is characterized by a quintuple $(\mathbf{X}, A, R, P, \gamma)$, where $\mathbf{X}$ is a finite set of states; $A$ is a finite set of actions; $R : \mathbf{X} \times A \rightarrow \mathbb{R}$ is the reward function of the agent; $P : \mathbf{X} \times A \times \mathbf{X} \rightarrow [0,1]$ is the transition function; and finally, $\gamma \in [0, 1)$ is the discount rate on future rewards. A factored-state Markov decision process (fMDP) is a structured MDP, where $\mathbf{X}$ is the Cartesian product of $m$ smaller components: $\mathbf{X} = X_1 \times X_2 \times \ldots \times X_m$. A function $f$ is a *local-scope* function if it is defined over a subspace $\mathbf{X}[Z]$ of the state space, where $Z$ is a (presumably small) index set. We make the standard assumption [9] that for each $i$ there exist sets $\Gamma_i$ of size $O(1)$ such that $\vec{x}_{t+1}[i]$ depends only on $\vec{x}_t[\Gamma_i]$ and $a_t$. The transition probabilities are then

**Algorithm 2** Reward learning in fMDPs with optimistic initialization

    **input:** $R_0$
    $\mathcal{M}^{-r} = \left(\{X_i\}_1^m; A; \{P_i\}_1^m; \vec{x}_s; \gamma; \{\Gamma_i\}_1^m; \{Z_j\}_1^J\right)$
    $t := 0, \vec{x}_0 := \vec{x}_s, m := 0, Q := I, \vec{w} := R_0\vec{1}$
    **repeat**
        $\widehat{\vec{r}} := Q\vec{w}$
        $a_t :=$ greedy action in fMDP($\mathcal{M}^{-r}, \widehat{\vec{r}}$), state $\vec{x}_t$
        execute $a_t$, observe reward $r_t$, next state $\vec{x}_{t+1}$
        $\vec{u} := \chi(\vec{x}_t, a_t)$
        $Q := Q - \frac{(Q\vec{u})(Q\vec{u})^T}{1+\vec{u}^T Q\vec{u}}, \vec{w} := \vec{w} + \vec{u}^T r_t$
        $t := t + 1$
    **until** there are more samples

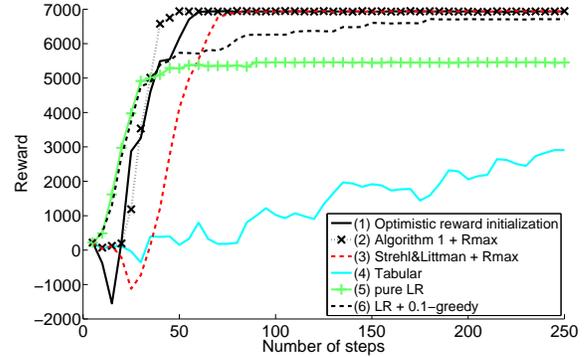

Figure 1: The value of the learned policy as a function of time for different reward-learning algorithms.

$P(\vec{y} \mid \vec{x}, a) = \prod_{i=1}^{m} P_i(\vec{y}[i] \mid \vec{x}[\Gamma_i], a)$ for each $\vec{x}, \vec{y} \in \mathbf{X}$, $a \in A$, so each factor is determined by a local-scope function. We make the analogous assumption that the reward function is the sum of $J$ local-scope functions with scopes $Z_j$: $R(\vec{x}, a) = \sum_{j=1}^{J} R_j(\vec{x}[Z_j], a)$. An fMDP is fully characterized by the tuple $\mathcal{M} = \left(\{X_i\}_{i=1}^m; A; \{P_i\}_{i=1}^m; \{R_j\}_{j=1}^J; \vec{x}_s; \gamma; \{\Gamma_i\}_{i=1}^m; \{Z_j\}_{j=1}^J\right)$. Algorithms exist for learning transition probabilities [9] and dependency structure [14], but until now, no algorithm existed for learning the reward functions.

For $J > 1$, we can only observe the *sums* of unknown quantities, not the quantities themselves. Doing so requires the solution of an online linear regression problem with a suitable encoding of the reward model. Let $N_r$ be the total number of parameters describing the reward model ($N_r \leq J|A|n^{n_f}$), and consider the indicator function $\chi(\vec{x}, a) \in \mathbb{R}^{N_r}$, which is 1 on indices corresponding to $R_j(\vec{x}[Z_j], a)$ and 0 elsewhere.

Our solution, Algorithm 2, is a modification of Algorithm 1. We initialize all unknown rewards to some constant $R_0$ (analogous to the common maximum reward parameter $R_{\max}$ [4]). If $R_0$ is sufficiently high, the algorithm outputs optimistic reward estimates for unknown states (instead of $\bot$), and otherwise gives near-accurate predictions with high probability. This property follows from standard arguments [17]: for unknown states, the noise term of the prediction error can be bounded by Azuma's inequality, and $R_0$ can be set high enough so that the second term is positive and dominates. This form of optimistic initialization has proven consistently better than R-max in flat MDPs [17]. For known states, the KWIK guarantees suffice to ensure near-optimal behavior. Because it combines a KWIK model-learner with R-max style exploration, this algorithm is PAC-compact-MDP—the first efficient algorithm for this task. In Section 5.1 we combine this algorithm with another KWIK learner that learns the transition dependency structure and probabilities [11] to learn the full fMDP model.

### 3.1 Experiments

We carried out reward-learning experiments on the Stocks domain [14], with 3 sectors and 2 stocks per sector. Rewards were uniformly random in the interval $[0.5, 1.5]$ for owning a rising stock, and random in $[-1.5, -0.5]$ for owning a decreasing stock. We compared six algorithms: (1) Algorithm 2; (2) Algorithm 1 modified to output $R_{\max}$ in unknown states; (3) the previous state-of-the-art KWIK-LR algorithm [15] modified to output $R_{\max}$ in unknown states; (4) a flat tabular reward learner; and to demonstrate the need for efficient exploration, (5) linear regression without exploration and (6) linear regression with epsilon-greedy exploration.

Each algorithm was run 20 times for 250 steps, updating the model every 5 steps. For the Stocks(3,2) domain, $R_{\max} = 6$. We used $R_0 = 10$ for Algorithm 2, $\alpha_0 = 1.0$ for the second approach and $\alpha_1 = \alpha_2 = 1.0$ for the third one. The values of the learned policies are shown in Figure 1. All curves except (1) and (2) differ significantly on a 95% confidence level. Notice (3) and (5) take longer to learn the model, (4) takes far longer, and (6) fails to explore and find the correct model.

## 4 Learning Transition Probabilities

We consider another novel application of KWIK linear regression—learning action-effect probabilities in environments where these effects may be ambiguous. Specifically, we consider environments where actions are encoded as stochastic *action schemas* (e.g. travel(X, Y) rather than travel(Paris, Rome)) and the effects of these actions are stochastic. For instance, the action travel(X, Y) may result in the effect $at$(Y) with probability .9 and the effect $at$(X) with probability .1. More formally, every action $a \in A$ is of the form $a = [(\omega_0, p_0) \cdots (\omega_n, p_n)]$ where each $\omega_i \in \Omega^a$ is a possible effect. When the action is taken, one of these effects occurs according to the probability distribution induced by the $p_i$s. The schemas may also contain conditional effect distributions, the nature of which is determined by the specific language used (as discussed in the following subsections). This form of generalization has been used to encode many different types of environments in RL, including stochastic STRIPS [12], Object Oriented MDPs (OOMDPs) [5], and typed dynamics [13]. Learning these probabilities is non-trivial because for a given state action pair ($s$, $a$), the effects are partitioned into equivalence classes $E(s, a) = \{\{\omega_i, \omega_j, \omega_k\}, \{\omega_l, \omega_m\}, ...\}$ where each $e \in E(s, a)$ contains effects that are identical given state $s$. For instance, if an action has two possible effects $Holding(X)$ and $Holding(Y)$, but is taken in a state where X and Y are both already held, we cannot tell which actually occurred. Notice that the probability of any equivalence class is equal to the sum of the probabilities of the effects it contains, hence the link to linear regression.

The standard "counting" method for learning probabilities cannot be used in such a setting because it is unclear which effect's count ($Holding(X)$s or $Holding(Y)$s) we should increment. However, KWIK-LR can be used to learn the probabilities. While standard linear regression could also be used if transition data was available in batch form, an RL agent using KWIK-LR can learn the probabilities online and maintain the PAC-compact-MDP guarantee needed for effective exploration. Algorithm 3 presents a PAC-compact-MDP algorithm for an agent that is given full action-operator specifications, except for the probabilities. That is, since our focus is on learning the probabilities, we assume that each $\omega_i$ is known, but not the $p_i$s. We also assume the preconditions and reward structure of the problem are known. We discuss methods for relaxing these assumptions in Section 5. This algorithm can be used with several representations including those mentioned above.

Intuitively, Algorithm 3 computes an optimistically optimal action $a_t$ using a planner (Line 7, detailed below) and then gathers experience indicating which of the equivalence classes $e_t \in E(s_t, a_t)$ actually occurred. For instance, it may see that $Holding(X)$ occurred, instead of $Holding(Y)$. This equivalence class contains one or more effects ($\omega_i, \omega_j... \in \Omega^a$), and an indicator vector $\vec{x}$ is created where $\vec{x}_i = 1$ if $\omega_i \in e_t$ (Line 10). For instance, if the agent was not holding anything, and then $Holding(X)$ occurred and not $Holding(Y)$, the vector would be $[1; 0]$, but in a state where both were already held, $\vec{x}$ would always come out $[1; 1]$. Note that each equivalence class in $E(s, a)$

---

**Algorithm 3** Transition Probability Learner

1: **input:** $S$, $A$ (action schemas sans $p_i$s), $R$, $\alpha$
2: $\forall a \in A$, instantiate a learner (Algorithm 1) $\mathcal{L}^a(\alpha)$
3: **for** each current state $s_t$ **do**
4:     **for** each $e \in E(s, a)$, $s \in S$ and $a \in A$ **do**
5:         Construct $\vec{x}$ where $x_j = 1$ if $j \in e$, else 0.
6:         $\widehat{P}(e) = $ Prediction of $\mathcal{L}^a(\vec{x})$ // can be $\perp$
7:     Perform modified value iteration on $\{S, A, E, \widehat{P}, R\}$, to get greedy policy $\pi$
8:     Perform action $a_t = \pi(s_t)$, observe $e_t \in E(s, a_t)$
9:     **for** equivalence classes $e \in E(s_t, a_t)$ **do**
10:        Construct $\vec{x}$ where $x_j = 1$ if $j \in e$, else 0.
11:        Update $\mathcal{L}^{a_t}$ with $\vec{x}$ and $y = 1$ if $e = e_t$, else $y = 0$

---

induces a unique (and disjoint) $\vec{x}$. *Each* of these is used to update the KWIK-LR learner (Lines 9-11), with an output ($y$) of 1 associated with the $\vec{x}$ that actually happened (which may have 1s in multiple places, as in the ambiguous *Holding* case). Given any possible action and equivalence class in the state/action space, the KWIK-LR agent can now be queried to determine the probability of the equivalent transition (Lines 5-6), though it may return $\perp$, identifying transitions from equivalence partitions that have unlearned probability.

Planning in Algorithm 3 is done by constructing a transition model in the grounded state space (Lines 4-6). KWIK-LR determines for each $E(s, a)$, the probability of the possible next states (one for each $e \in E(s, a)$). A modified version of value iteration [16] (Line 7) is then used to plan the optimal next action. The modification is that at every iteration, for every state-action pair that has effects with known probabilities $K = \{\omega_i, \omega_j...\}$ and unknown probabilities $U = \{\omega_k, \omega_l...\}$, the effect in $U$ that leads to the highest value next state is considered to have probability $1 - \sum_{\omega_i \in K} P(\omega_i)$. This change is designed to force value iteration to be optimistic—it considers the most rosiest of all models consistent with what has been learned. We note that planning in the flat state space can require exponential computation time, but this is often unavoidable, and since the model is still *learned* in a sample efficient manner, it satisfies the conditions for PAC-compact-MDP.

### 4.1 Application 2: Stochastic STRIPS

STRIPS domains [8] are made up of a set of objects $\mathcal{O}$, a set of predicates $P$, and a set of actions $A$. The actions have conjunctive (over $P$) preconditions and effects specified by **ADD** and DELETE (**DEL**) lists, which specify what predicates are added and deleted from the world state when the action occurs. Stochastic STRIPS operators generalize this representation

> $paint(X)$: reward $= -1$
> **PRE:** none
> **ADD**: $Painted(X)$ **DEL:** none **0.6**
> **ADD**: $Painted(X), Scratched(X)$ **DEL:** none **0.3**
> **ADD**: none **DEL:** none **0.1**
>
> $polish(X)$: reward $= -1$
> **PRE:** none
> **ADD**: none **DEL:** $Painted(X)$ **0.2**
> **ADD**: none **DEL:** $Scratched(X)$ **0.2**
> **ADD**: $Polished(X)$ **DEL:** $Painted(X), Scratched(X)$ **0.3**
> **ADD**: $Polished(X)$ **DEL:** $Painted(X)$ **0.2**
> **ADD**: none **DEL:** none **0.1**
>
> $shortcut(X)$: reward $= -1$
> **PRE:** none
> **ADD**: $Painted(X), Polished(X)$ **DEL:** none **0.05**
> **ADD**: none **DEL:** none **0.95**
>
> $done(X)$: reward $= 10$
> **PRE:** $Painted(X), Polished(X), Scratched(X)$
> **ADD**: $Finished(X)$ **DEL:** none **1.0**

Table 1: Stochastic Paint/Polish world

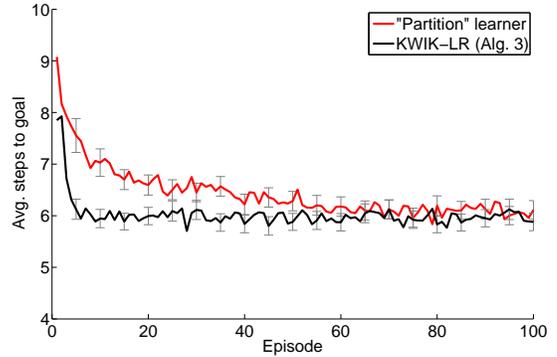

Figure 2: Average steps to the goal over 1000 runs.

### 4.2 Application 3: Stochastic Object Oriented RL

Object-oriented MDPs [5] consist of a set of objects $\mathcal{O}$, a set of actions $A$ that take elements of $\mathcal{O}$ as parameters, and (in the original deterministic description) a set of condition-effect pairs $\langle c, \omega \rangle$ associated with each action. Objects have attributes, and the set of all objects and their attribute values at a given time $t$ constitute the state $s_t$. When an action $a$ is executed from $s_t$, the environment checks which condition $c_i \in C$ is satisfied (if any), and applies the corresponding effect $w_i \in \Omega^a$, which updates the attributes of the affected objects. Stochastic OOMDP effects generalize this representation so that a given condition induces not just a single effect ($\omega$) but a distribution over possible effects ($\Omega^{ca}$), only one of which ($\omega_i$) will actually occur on a single timestep. In the parlance of action schemas as defined above, each $\{c, \Omega\}$ pair for action $a$ can be thought of as its own action with preconditions specified by $c$. Again, this model can be viewed as a specific version of the action schemas presented above.

by considering multiple possible action effects specified by $\langle \mathbf{ADD}, \mathbf{DEL}, \mathbf{PROB} \rangle$ tuples as in Table 1. Notice this representation is an instance of the general action schemas defined above. While others have looked at learning similar operators [12], their work attempted to heuristically learn the full operators (including structure), and could not give any guarantees (as we do) on the behavior of their algorithm, nor did they identify an efficient algorithm for learning the probabilities, as we have with Algorithm 3. To make the planning step well defined, we consider Stochastic STRIPS with *rewards* [18].

Table 1 illustrates a variant of the familiar Paint/Polish domain in the Stochastic STRIPS setting. There are several ambiguous effects. For instance, executing *paint* on a scratched but not painted object, and observing it is now scratched and painted, one cannot tell which of the first two effects occurred. We used this domain with a single object to empirically test Algorithm 3 against a version of the algorithm that does not use KWIK-LR, instead attempting to learn every possible equivalence class partition distribution separately (*Partition*). Because the focus here is on learning the transition probabilities ($p_i$s), both learners were given the preconditions and effects ($\omega_i$) of each action. We discuss relaxing these assumptions later. Figure 2 shows the results (with the known/unknown thresholds empirically tuned) averaged over 1000 runs with randomized initial states for each episode (both learners receive the same initial states). Algorithm 3 learns much faster because it effectively shares information between partitions.

Previous work [5] presented an efficient algorithm for learning deterministic effects. Here, we demonstrate Algorithm 3 learning the probabilities associated with each effect in the stochastic setting when the possible effects given a condition and the conditions themselves are known in advance. Methods for relaxing these assumptions are discussed in later sections.

We demonstrate Algorithm 3 for stochastic OOMDPs on a simple $5 \times 5$ Maze domain, illustrated in Figure 3. The agent starts at location $S$ and the goal is to arrive at $G$. Each step has a cost of $-0.01$, and arriving at the goal results in a reward of $+1$. The agent's actions are N, S, E and W. When executing an action, the agent will attempt to move in the desired direction with probability 0.8 and will slip to either side with probability 0.1. If it hits a wall, it stays put. This rule is what produces ambiguity in the effects. For example, imagine the agent has a wall to its North and East. If it attempts the N action, it could move to the West (with probability 0.1), or stay in place. If it stays in place, it might be because it attempted to move North (with probability 0.8) and hit the North wall, or it attempted to move East (with probability 0.1) and hit the East wall.

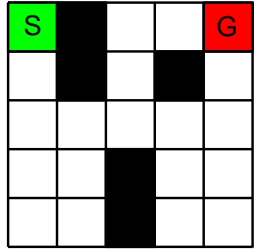

Figure 3: Maze domain.

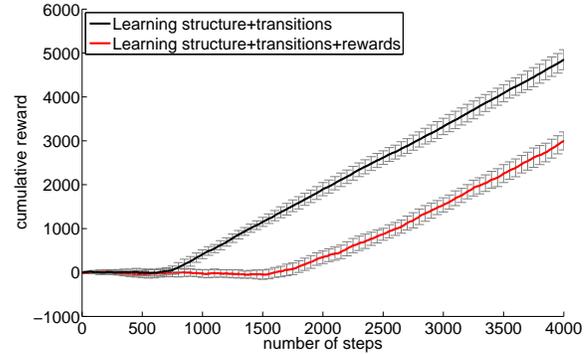

Figure 5: Average cumulative rewards over 10 runs on the Stocks domain.

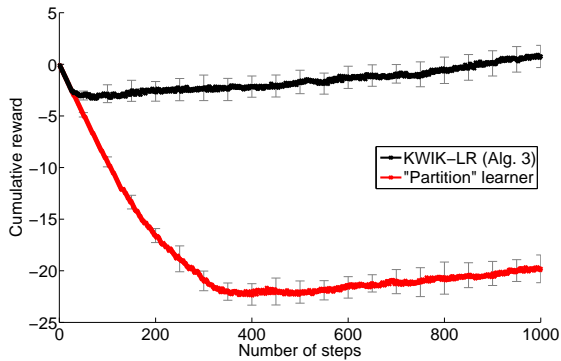

Figure 4: Stochastic OORL results for KWIK-LR and Partition in the Maze domain.

Figure 4 presents the results. As in the previous experiment, we can see that the KWIK-RL algorithm learns much faster than the "Partition" learner by sharing information between equivalence partitions.

## 5 Extensions

We now discuss extensions for learning more of the compact representations discussed in this work. We will outline how to learn full fMDPs and larger parts of STRIPS and OOMDP models by combining KWIK-LR with other KWIK learners, and provide empirical support in the fMDP case. We also discuss a variation of KWIK-LR that can be used to learn stochastic action-schema outcomes ($\Omega^a$).

### 5.1 Combining with Other KWIK Learners

Existing work [11] describes methods for combining simple KWIK agents to learn in increasingly complex domains. We follow this "building blocks" approach and show how to combine the fMDP reward-learning algorithm (Algorithm 2) with an implementation of the Noisy-Union [11; 10] algorithm (which is also KWIK) to learn the transition structure ($\Gamma$), transition probabilities ($P$), and reward function ($R$) of an fMDP all at once. The only knowledge given to the agent is the number of parents a factor may have ($|Z|$) and the reward-function structure, resulting in the only algorithm to date that learns all of these parameters efficiently. Experience is fed to both building block algorithms in parallel. The reward learner outputs an optimistic approximation of the reward function, which is given to Noisy-Union, which then learns the transition structure and probabilities. We conducted experiments to validate this algorithm in the Stocks domain. For comparison, we also ran the Noisy-Union algorithm with the rewards given *a priori*. Figure 5 displays the results, which show that all three quantities can be learned with small overhead.

We note that there are many other settings that could benefit from combining KWIK-LR with other KWIK learners for different parts of the model. For instance, in stochastic STRIPS and OOMDPs, the preconditions of actions (STRIPS) or the conditional effects (OOMDP) can be learned using an existing KWIK adaptation of Noisy Union [6] as long as their size is bounded by some known constant. Together, these learners could be used to learn all but the effects ($\Omega^a$) of each action operator, a problem we now consider.

### 5.2 Future Work: Learning Effects

All of the effect distribution / condition learning variations require the possible effects ($\Omega^a$) as input. Unfortunately, relaxing this assumption in the stochastic case is unlikely, since the effect learning problem is known to be NP-Hard [12]. When the number of possible effects is very small, or each $\omega_i$ is of constant size, enumeration techniques could be used. But, these assumptions are often violated, so researchers have concentrated on heuristic solutions [12]. Here, we suggest a novel heuristic that extends KWIK-LR for probability learning to the setting where the whole action schema (including $\Omega^a$) needs to be learned.

We propose using sparsification; we consider all possible effects and use KWIK-LR to learn their probabilities, with an extra constraint that the number of "active" (non-zero) probabilities should be small. This minimization can be efficiently computed via linear programming, and techniques such as column generation may be used to keep the number of active constraints small in most cases. Together with the other learners discussed in this paper, the implementation of such an extension would form a complete solution to the action-schema learning problem: the probabilities and preconditions can be learned with KWIK-learners and the effects themselves can be learned heuristically using this sparsification extension. It remains a matter for future work to compare this system to other heuristic solutions [12] for such problems.

## 6 Related Work

Online linear regression has also been studied in the regret minimization framework (see e.g. [1]). Applications to restricted RL problems also exist [2], but with different types of bounds. Furthermore, regret analysis seems to lack the modularity (the ability to combine different learners) of the KWIK framework.

Previous work on linear regression for model-based RL has focussed on learning linear transition functions in continuous spaces. However, these approaches often lacked theoretical guarantees or placed restrictions on the environment regarding noise and the reward structure [7]. In this paper, we have both improved on the current state-of-the-art algorithm for linear regression in RL [15], and used it in applications beyond the standard linear transition models. Our theoretical results and experiments illustrate the potential of KWIK-LR in model-based RL. In the future, we intend to identify other compact models where this technique can facilitate efficient learning and perform expanded empirical and theoretical studies of the extensions mentioned above.

**Acknowledgements**

We thank Alexander L. Strehl and the reviewers for helpful comments. Support was provided by the Fullbright Foundation, DARPA IPTO FA8750-05-2-0249, FA8650-06-C-7606, and NSF IIS-0713435.